\documentclass[11pt]{article}
\usepackage{eacl2017}
\usepackage{times}
\usepackage{url}
\usepackage{latexsym}

\usepackage{amsmath}
\usepackage{amssymb}
\usepackage{stmaryrd}
\usepackage{array}
\usepackage{graphicx}
\usepackage{cases}
\usepackage{watermark}
\usepackage{pbox}
\usepackage{multirow}
\usepackage{soul}
\setulcolor{red}
\usepackage{enumitem}
\usepackage{xspace}
\usepackage{makecell}
\usepackage[draft,textsize=footnotesize]{todonotes} 
\usepackage[ruled,noline,linesnumbered]{algorithm2e}
\usepackage{xspace}
\usepackage[small]{caption}
\usepackage{subcaption}
\usepackage{anyfontsize}
\usepackage{xcolor}

\newcommand{\corpuslong}{\textbf{J}HU \textbf{FL}uency-\textbf{E}xtended \textbf{G}{\sc ug} corpus\xspace}
\newcommand{\corpus}{{\sc jfleg}\xspace}

\newcommand{\gug}{{\sc gug}\xspace}
\newcommand{\gleu}{{\sc gleu}\xspace}
\newcommand{\mm}{{\sc M$^2$}\xspace}
\newcommand{\camb}{{\sc camb\small{14}}\xspace}
\newcommand{\nus}{{\sc nus}\xspace}
\newcommand{\nmt}{{\sc camb\small{16}}\xspace}
\newcommand{\amu}{{\sc amu}\xspace}
\newcommand{\nucle}{{\sc nucle}\xspace}
\newcommand{\fce}{{\sc fce}\xspace}
\newcommand{\aesw}{{\sc aesw}\xspace}
\newcommand{\gec}{{\sc gec}\xspace}

\newcommand{\mt}{{\sc mt}\xspace}

\newcommand\fauxsc[1]{\fauxschelper#1 \relax\relax}%
\def\fauxschelper#1 #2\relax{%
    \fauxschelphelp#1\relax\relax%
    \if\relax#2\relax\else\ \fauxschelper#2\relax\fi%
}
\def\Hscale{.85}\def\Vscale{.74}\def\Cscale{1.12}%
\def\fauxschelphelp#1#2\relax{%
    \ifnum`#1>``\ifnum`#1<`\{\scalebox{\Hscale}[\Vscale]{\uppercase{#1}}\else%
    \scalebox{\Cscale}[1]{#1}\fi\else\scalebox{\Cscale}[1]{#1}\fi%
    \ifx\relax#2\relax\else\fauxschelphelp#2\relax\fi}%

\eaclfinalcopy

\title{
JFLEG: A Fluency Corpus and Benchmark for Grammatical Error Correction
}
\newcommand{\clsp}{\ensuremath{{}^\text{1}}}
\newcommand{\grammarly}{\ensuremath{{}^\text{2}}}

\author{
  Courtney Napoles\textnormal{,\clsp}
   Keisuke Sakaguchi\textnormal{,\clsp} \and
   Joel Tetreault\textnormal{\grammarly}\\
 \clsp Center for Language and Speech Processing, Johns Hopkins University \\
 \grammarly Grammarly\\
 \tt{\{napoles,keisuke\}@cs.jhu.edu},
 \url{joel.tetreault@grammarly.com}}

\date{}

\begin{document}
\maketitle
\begin{abstract}
We present a new parallel corpus, \corpuslong (\corpus) 
for developing and evaluating grammatical error correction (\gec).
Unlike other corpora, it represents a broad range of language proficiency levels and uses holistic \textit{fluency edits} to not only correct grammatical errors but also make the original text more native sounding.
We describe the types of corrections made and benchmark four leading \gec systems on this corpus,
identifying specific areas in which they do well and how they can improve.
\corpus fulfills the need for a new gold standard to properly assess the current state of \gec.

\end{abstract}

\section{Introduction}
\label{sec:intro}
Automatic grammatical error correction (\gec) progress is limited by the corpora available for developing and evaluating systems.
Following the release of the test set of the CoNLL--2014 Shared Task on \gec \cite{ng-EtAl:2014:W14-17}, systems have been compared and new evaluation techniques proposed on this single dataset. 
This corpus has enabled substantial advancement in \gec beyond the shared tasks, but we are concerned that the field is over-developing on this dataset.
This is problematic for two reasons: 1) it represents one specific population of language learners; and 2) the corpus only contains \textit{minimal edits}, which correct the grammaticality of a sentence but do not necessarily make it \textit{fluent} or native-sounding.

\begin{table}
    \fontsize{10}{12}\selectfont
\begin{tabular}{|p{7.2cm}|}
\hline 
\textbf{Original:} they just creat impression such well that people are drag to buy it .\\ \hline
\textbf{Minimal edit:} They just create an impression so well that people are dragged to buy it .\\ \hline
\textbf{Fluency edit:} They just create such a good impression that people are compelled to buy it.\\ \hline
\end{tabular}

\caption{\label{tab:example}  A sentence corrected with just minimal edits compared to fluency edits.}
\vspace{-2mm}
\end{table} 

To illustrate the need for fluency edits, consider the example in Table \ref{tab:example}.  The correction with only minimal edits is grammatical but sounds \textit{awkward} (unnatural to native speakers). The fluency correction has more extensive changes beyond addressing grammaticality, and the resulting sentence sounds more natural and its intended meaning is more clear.
It is not unrealistic to expect these changes from automatic \gec: the current best systems use machine translation (\mt) and are therefore capable of making broader sentential rewrites but, until now, there has not been a gold standard against which they could be evaluated.

Following the recommendations of \newcite{sakaguchi2016reassessing}, we release a new corpus for \gec, the \corpuslong (\corpus), which adds a layer of annotation to the \gug corpus \cite{heilman-EtAl:2014:P14-2}. \gug represents a cross-section of ungrammatical data, containing sentences written by English language learners with different L1s and proficiency levels. 
For each of 1,511 \gug sentences, we have collected four human-written corrections which contain holistic fluency rewrites instead of just minimal edits.
This corpus represents the diversity of edits that \gec needs to handle and sets a gold standard to which the field should aim. 
We overview the current state of \gec by evaluating the performance of four leading systems on this new dataset. 
We analyze the edits made in \corpus and summarize which types of changes the systems successfully make, and which they need to address.
\corpus will enable the field to move beyond minimal error corrections.

\section{\fauxsc{gec} corpora}\label{sec:corpora}
\vspace{-2mm}
There are four publicly available corpora of non-native English annotated with corrections, 
to our knowledge.
The {\sc nus} Corpus of Learner English (\nucle) contains essays written by students at the National University of Singapore, corrected by two annotators using 27 error codes \cite{dahlmeier-ng-wu:2013:BEA8}. 
The CoNLL Shared Tasks used this data \cite{ng-EtAl:2014:W14-17,ng-EtAl:2013:CoNLLST}, and the 1,312 sentence test set from the 2014 task has become \textit{de rigueur} for benchmarking \gec.  This test set has been augmented with ten additional annotations from Bryant et al. \shortcite{bryant-ng:2015:ACL-IJCNLP} and eight from Sakaguchi et al. \shortcite{sakaguchi2016reassessing}. 
The Cambridge Learner Corpus First Certificate in English (\fce) has essays coded by one rater using about 80 error types, alongside the score and demographic information \cite{yannakoudakis-briscoe-medlock:2011:ACL-HLT2011}. 
The Lang-8 corpus of learner English is the largest, with text from the social platform lang-8.com automatically aligned to user-provided corrections \cite{tajiri-komachi-matsumoto:2012:ACL2012short}. 
Unlimited annotations are allowed per sentence, but 87\% were corrected once and 12\% twice.
The \aesw 2016 Shared Task corpus contains text from scientific journals corrected by a single editor. 
To our knowledge, \aesw is the only corpus that has not been used to develop a \gec system.

We consider \nucle\footnote{Not including the additional fluency edits collected for the CoNLL-2014 test set by \newcite{sakaguchi2016reassessing}.} and \fce to contain \textit{minimal edits}, since the edits were constrained by error codes, and the others to contain \textit{fluency} edits since there were no such restrictions.
English proficiency levels vary across corpora: \fce and \nucle texts were written by English language learners with relatively high proficiency, but Lang-8 is open to any internet user.
\aesw has technical writing by the most highly proficient English writers. 
Roughly the same percent of sentences from each corpus is corrected, except for \fce which has significantly more. This may be due to the rigor of the annotators and not the writing quality.

\begin{table}[t]
 	\centering
 	\small
 	\begin{tabular}{l|c|c|c|c}
    \hline
 		& & \textbf{Mean chars}	& \textbf{Sents.}	& \textbf{Mean}\\
 		\textbf{Corpus}	& \textbf{\# sents.}	& \textbf{per sent.}	& \textbf{changed} 	& \textbf{LD} \\
 		\hline
 		\aesw     & 1.2M	& 133	& 39\%  & 3\\
 		\fce	  & 34k		& 74	& 62\%  & 6	\\
 		Lang-8    & 1M		& 56	& 35\% 	& 4 \\
 		\nucle    & 57k		& 115	& 38\% 	& 6\\
 		\hline
 		\corpus   & 1,511	& 94	& 86\%	& 13 \\
        \hline
 	 \end{tabular}
\caption{\label{tab:corpora} Parallel corpora available for \gec.} 
\vspace{-2mm}
 \end{table}
 
The following section introduces the \corpus corpus, which represents a diversity of potential corrections with four corrections of each sentence.
Unlike \nucle and \fce, \corpus does not restrict corrections to minimal error spans, nor are the errors coded. Instead, it contains holistic sentence rewrites, similar to Lang-8 and \aesw, but contains more reliable corrections than Lang-8 due to perfect alignments and screened editors, and more extensive corrections than \aesw, which contains fewer edits than the other corpora with a mean Levenshtein distance (LD) of 3 characters. Table \ref{tab:corpora} provides descriptive statistics for the available corpora.
\corpus is also the only corpus that provides corrections alongside sentence-level grammaticality scores of the uncorrected text.

\section{The \fauxsc{jfleg} corpus}\label{sec:data}
\vspace{-2mm}
Our goal in this work is to create a corpus of fluency edits, following the recommendations of \cite{sakaguchi2016reassessing}, who identify the shortfalls of minimal edits: they artificially restrict the types of changes that can be made to a sentence and do not reflect the types of changes required for native speakers to find sentences \textit{fluent}, or natural sounding.
We collected annotations on a public corpus of ungrammatical text, the \gug (Grammatical/Ungrammatical) corpus \cite{heilman-EtAl:2014:P14-2}.
\gug contains $3.1k$ sentences written by English language learners for the 
TOEFL$^\textrm{\textregistered}$ exam, covering a range of topics. 
The original \gug corpus is annotated with grammaticality judgments for each sentence, ranging from 1--4, where 4 is perfect or native sounding, and 1 incomprehensible. The sentences were coded by five crowdsourced workers and one expert.
We refer to the mean grammaticality judgment of each sentence from the original corpus as the \textit{\fauxsc{gug} score}.

In our extension, \corpus, the 1,511 sentences which comprise the \gug development and test sets were corrected four times each on Amazon Mechanical Turk.
Annotation instructions are included in Table \ref{tab:instructions}.
50 participants from the United States 
passed a qualifying task of correcting five sentences, which was reviewed by the authors (two native and one proficient non-native speakers of American English).
Annotators also rated how difficult it was for them to correct each sentence on a 5-level Likert scale (5 being very easy and 1 very difficult).
On average, the sentences were relatively facile to correct (mean difficulty of $3.5\pm 1.3$), which moderately correlates with the \gug score (Pearson's $r = 0.47$), 
indicating that less grammatical sentences were generally more difficult to correct.
To create a blind test set for the community, we withhold half (747) of the sentences from the analysis and evaluation herein.

\begin{table}[t]
\centering\small
\begin{tabular}{|p{7.1cm}|}
\hline
Please correct the following sentence to make it sound natural and fluent to a native speaker of (American) English. The sentence is written by a second language learner of English. You should fix grammatical mistakes, awkward phrases, spelling errors, etc. following standard written usage conventions, but your edits must be conservative. Please keep the original sentence (words, phrases, and structure) as much as possible. The ultimate goal of this task is to make the given sentence sound natural to native speakers of English without making unnecessary changes. Please do not split the original sentence into two or more. Edits are not required when the sentence is already grammatical and sounds natural.\\
\hline
\end{tabular}
\caption{\label{tab:instructions} \corpus annotation instructions.}
\end{table}

\begin{table}[t]
	\centering
	\small
\begin{tabular}{ccl|c|c|c}
	\hline
		& && \multicolumn{3}{c}{\textit{Error type in original}}\\
		& && \textbf{Awkward}	& \textbf{Ortho.}	& \textbf{Grammatical} \\
	\hline
\rule{0pt}{2.2ex}\parbox[t]{2mm}{\multirow{2}{*}{\rotatebox[origin=c]{90}{\textit{Edit}}}} &
\hspace{-4mm} \parbox[t]{2mm}{\multirow{2}{*}{\rotatebox[origin=c]{90}{\textit{type}}}}
 & \hspace{-2mm}Fluency		& 38\%		& 35\%		& 32\% \\
 & &\hspace{-2mm}Minimal 		& 82\%		& 89\%		& 85\% \\
 \hline
\end{tabular}
\caption{\label{tab:orig-src-edits} Percent of sentences by error type that were changed with fluency or minimal edits.}
\vspace{-1mm}
\end{table}

The mean LD between the original and corrected sentences is more than twice that of existing corpora (Table \ref{tab:corpora}).
LD negatively correlates with the \gug score ($r = -0.41$) and the annotation difficulty score ($-0.37$), supporting the intuition that less grammatical sentences require more extensive changes, and it is harder to make corrections involving more substantive edits.
Because there is no clear way to quantify agreement between annotators, we compare the annotations of each sentence to each other.
The mean LD between all pairs of annotations is greater than the mean LD between the original and corrected sentences (15 characters), however 36\% of the sentences were corrected identically by at least two participants. 

Next, the English L1 authors examined 100 randomly selected original and human-corrected sentence pairs and labeled them with the type of error present in the sentence and the type of edit(s) applied in the correction.
The three error types are sounds \textit{awkward} or has an \textit{orthographic} or \textit{grammatical} error.\footnote{Due to their frequency, we separate orthographic errors (spelling and capitalization) from other grammatical errors.} 
The majority of the original sentences have at least one error (81\%), 
and, for 68\% of these sentences, the annotations are error free.
Few annotated sentences have orthographic (4\%) or grammatical (10\%) errors, but awkward errors are more frequent (23\% of annotations were labeled \textit{awkward})---which is not very surprising given how garbled some original sentences are and the dialectal variation of what sounds awkward.

The corrected sentences were also labeled with the type of changes made (minimal and/or fluency edits).  Minimal edits reflect a minor change to a small span (1--2 tokens) addressing an immediate grammatical error, such as number agreement, tense, or spelling. 
Fluency edits are more holistic and include reordering or rewriting a clause, and other changes that involve more than two contiguous tokens.  
69\% of annotations contain at least one minimal edit, 25\% a fluency edit, and 17\% both fluency and minimal edits.
The distribution of edit types is fairly uniform across the error type present in the original sentence (Table \ref{tab:orig-src-edits}).
Notably, fewer than half of awkward sentences were corrected with fluency edits, which may explain why so many of the corrections were still \textit{awkward}.

\section{Evaluation}

To assess the current state of \gec, we collected automated corrections of \corpus from four leading \gec systems with no modifications.
They take different approaches but all use some form of \mt.
The best system from the CoNLL-2014 Shared Task is a hybrid approach, combining a rule-based system with \mt and language-model reranking (\camb; Felice et al., 2014\nocite{felice-EtAl:2014:W14-17}).
Other systems have been released since then and report improvements on the 2014 Shared Task. 
They include a neural \mt model (\nmt; Yuan and Briscoe, 2016\nocite{yuan-briscoe:2016:N16-1}), a phrase-based \mt ({\sc pbmt}) with sparse features (\amu; Junczys-Dowmunt and Grundkiewicz, 2016\nocite{junczysdowmunt-grundkiewicz:2016:EMNLP2016}), and a hybrid system that incorporates a neural-net adaptation model into {\sc pbmt} (\nus; Chollampatt et al., 2016\nocite{chollampatt-hoang-ng:2016:EMNLP2016}).

We evaluate system output against  the four sets of \corpus corrections with \gleu, an automatic fluency metric specifically designed for this task \cite{napoles-EtAl:2015:ACL-IJCNLP} and the Max-Match metric (\mm) \cite{dahlmeier-ng:2012:NAACL-HLT}.
\gleu is based on the \mt metric {\sc bleu}, and represents the n-gram overlap of the output with the human-corrected sentences, penalizing n-grams that were been changed in the human corrections but left unchanged by a system.
It was developed to score fluency in addition to minimal edits since it does not require an alignment between the original and corrected sentences.
\mm was designed to score minimal edits and was used in the CoNLL 2013 and 2014 shared tasks on \gec \cite{ng-EtAl:2013:CoNLLST,ng-EtAl:2014:W14-17}. Its score is the F$_{0.5}$ measure of word and phrase-level changes calculated over a lattice of changes made between the aligned original and corrected sentences.
Since both \gleu and \mm have only been evaluated on the CoNLL-2014 test set, we additionally collected human rankings of the outputs to determine whether  human judgments of relative grammaticality agree with the metric scores when the reference sentences have fluency edits.

\begin{table}[!t]
    \centering
    \small
    \begin{tabular}{l|c|c|c||c}
    \hline
        & & & & \textbf{Sentences}\\ 
        \textbf{System} & \textbf{TrueSkill}	& \textbf{\fauxsc{gleu}} & \textbf{M}$^\textrm{\textbf{2}}$	& \textbf{changed}\\ 
        \hline
        Original	 & -1.64 			& 38.2 			&	0.0	& -- \\ 
        \hline
		{\sc camb}{\fontsize{8}{9.6}\selectfont 16}  		 & \textbf{0.21} 	& \textbf{47.2} & 50.8	& 74\% \\ 
        \sc{nus}	 & -0.20$^*$		& 46.3 			& \textbf{52.7}	& 69\% \\ 
        \sc{amu}     & -0.46$^*$		& 41.7 			& 43.2  & 56\% \\ 
		{\sc camb}{\fontsize{8}{9.6}\selectfont 14} 		 & -0.51$^*$		& 42.8 			& 46.6	& 58\% \\ 
		\hline 
		Human  		 & 2.60 			& 55.3 	& 63.2	& 86\%\\ 
        \hline
    \end{tabular}
\caption{\label{tab:gleu} Scores of system outputs. $^*$ indicates no significant difference from each other.}
\vspace{-1mm}
\end{table}

The two native English-speaking authors ranked six versions of each of 150 \corpus sentences: the four system outputs, one randomly selected human correction, and the original sentence.
The absolute human ranking of systems was inferred using TrueSkill, which computes a relative score from pairwise comparisons, and we cluster systems with overlapping ranges into equivalence classes by bootstrap resampling \cite{sakaguchi-post-vandurme:2014:W14-33,HerbrichMG06}.
The two best ranked systems judged by humans correspond to the two best \gleu systems, but \gleu switches the order of the bottom two.
The \mm ranking does not perform as well, reversing the order of the top two systems and the bottom two (Table \ref{tab:gleu}).\footnote{No conclusive recommendation about the best-suited metric for evaluating fluency corrections can be drawn from these results. With only four systems, there is no significant difference between the metric rankings, and even the human rank has no significant difference between three systems.}
The upper bound is \gleu = 55.3 and \mm = 63.2, the mean metric scores of each human correction compared to the other three.
\nmt and \nus are halfway to the gold-standard performance measured by \gleu and, according to \mm, they achieve approximately 80\% of the human performance.
The neural methods (\nmt and \nus) are substantially better than the other two according to both metrics. 
This ranking is in contrast to the ranking of systems on the CoNLL-14 shared task test against minimally edited references.
On these sentences, \amu, which was tuned to \mm, achieves the highest \mm score with 49.5 and \nmt, which was the best on the fluency corpus, ranks third with 39.9.

\begin{figure}[!t]
	\vspace{-2mm}
    \includegraphics[width=\columnwidth]{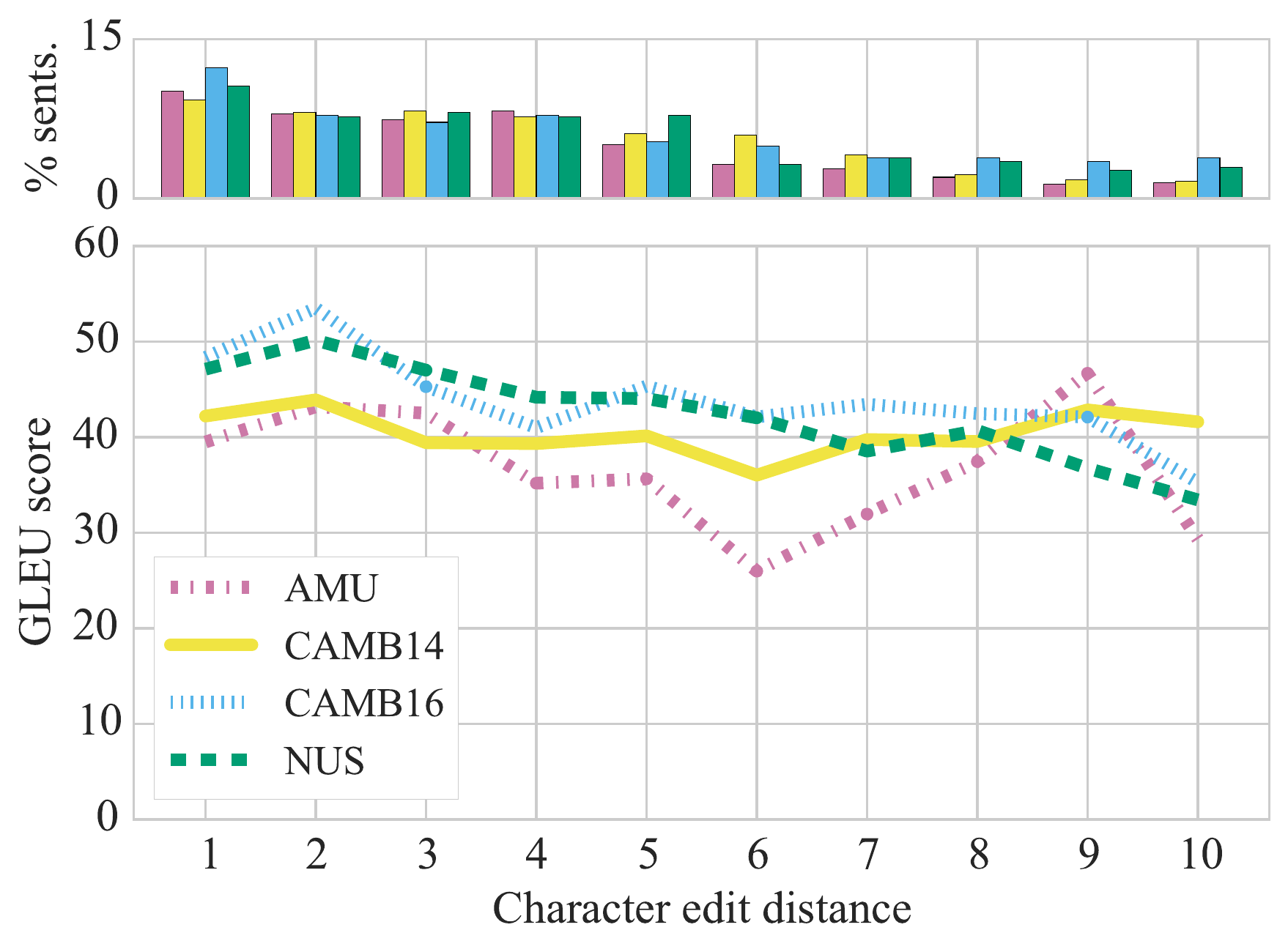}
   	\vspace{-8mm}
   \caption{\label{fig:gleu-ld} \gleu score of system output by LD from input.}
\vspace{-2mm}
\end{figure}

We find that the extent of changes made in the system output is negatively correlated to the quality as measured by \gleu (Figure \ref{fig:gleu-ld}). 
The neural systems have the highest scores for nearly all edit distances, and generate the most  sentences with higher LDs. \camb has the most consistent \gleu scores.
The \amu scores of sentences with LD $>6$ are erratic due to the small number of sentences it outputs with that extent of change.

\begin{figure}[t]
	\centering
    \includegraphics[width=7.5cm]{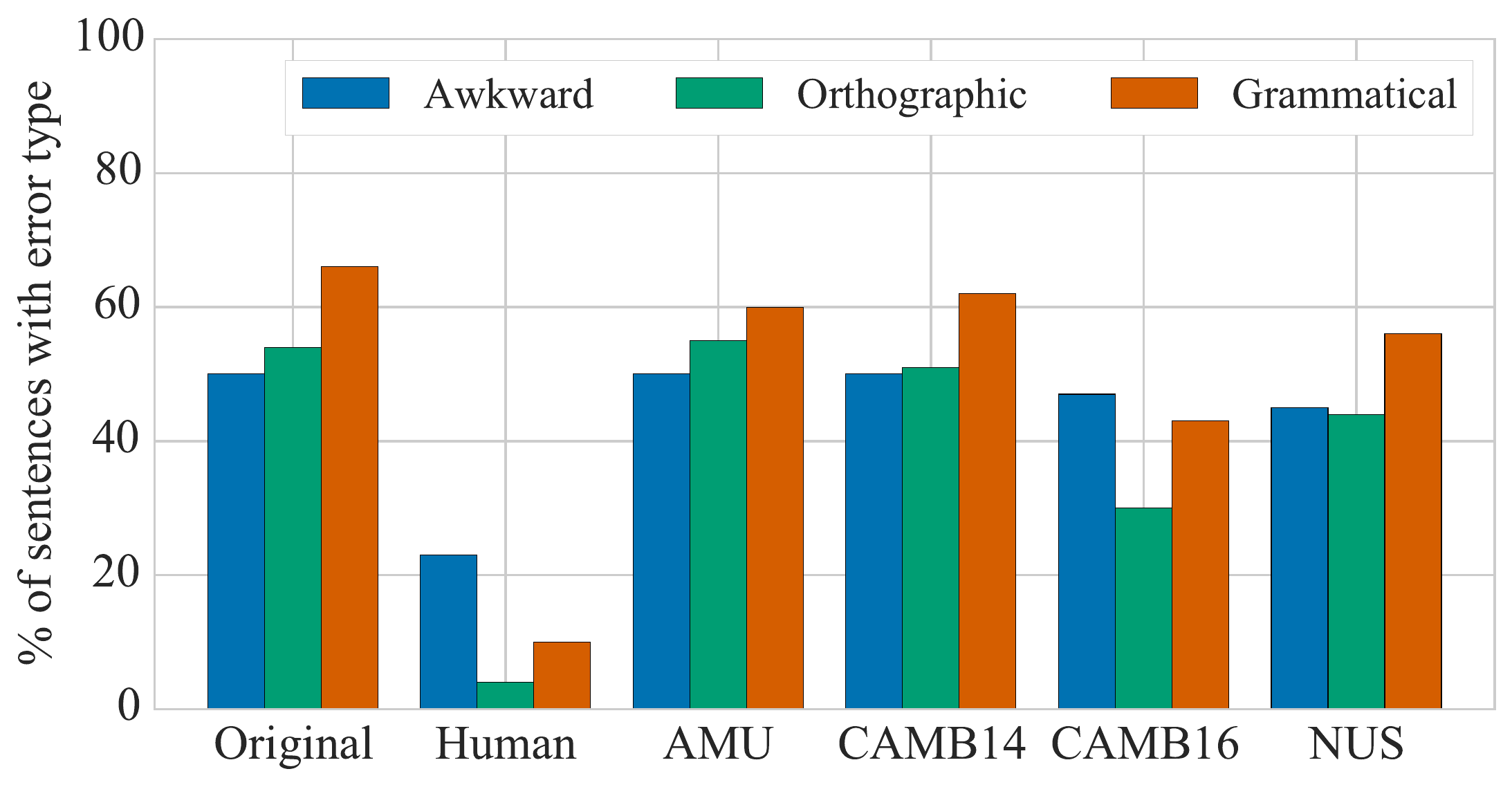}
    \vspace{-2mm}
	\caption{\label{fig:sys-analysis} Types of errors present in the original, annotated, and system output sentences.}
    
\end{figure}

\begin{table}[!t]
	\centering
	\small
	\begin{tabular}{cc|c|c|c}
		\hline
		& & \multicolumn{3}{c}{\textit{Error type in original}}\\
		& & \textbf{Awkward}	& \textbf{Ortho.}	& \textbf{Grammatical} \\
		\hline
		\multirow{2}{35pt}{\amu}\hspace{-2mm}	& F	& 2\%	& 2\%	& 2\%	\\
								& M	& 60\%	& 60\%	& 64\% \\
		\hline
		\multirow{2}{35pt}{{\sc camb}{\fontsize{8}{9.6}\selectfont 14}}\hspace{-2mm}	& F	& 2\%	& 0\%	& 2\% \\
								& M	& 64\%	&  69\%	& 65\% \\
		\hline
		\multirow{2}{35pt}{{\sc camb}{\fontsize{8}{9.6}\selectfont 16}}\hspace{-2mm}	& F	& 8\%	& 7\%	& 6\% \\
								& M	& 82\%	& 85\%	& 79\% \\
		\hline
		\multirow{2}{35pt}{\nus}\hspace{-2mm}	& F	& 4\%	& 4\%	& 3\% \\
								& M	& 68\%	& 81\%	& 79\% \\
		\hline
	\end{tabular}
	\caption{\label{tab:sys-src-edits} Percent of sentences by error type changed in system output with fluency (F) and minimal (M) edits.}
\vspace{-1mm}
\end{table}

\begin{table*}[!h]
\centering
\fontsize{9}{11}\selectfont
\begin{tabular}{|l|l|} 
\hline
\textbf{Original}	& First , advertissment make me to buy some thing unplanly .\\
\hline
\textbf{Human}		& First , an advertisement made me buy something unplanned .\\
\hline
\textbf{\fauxsc{amu}}		& First , advertissment makes me to buy some thing unplanly .\\
\hline
\textbf{\fauxsc{camb}{\scriptsize 14}}		& First , advertisement makes me to buy some things unplanly .\\
\hline
\textbf{\fauxsc{camb}{\scriptsize 16}}		& First , please let me buy something bad .\\
\hline
\textbf{\fauxsc{nus}}		& First , advertissment make me to buy some thing unplanly .\\
\hline
\hline
\textbf{Original}	& For example , in 2 0 0 6 world cup form Germany , as many conch wanna term work .	\\ \hline
\textbf{Human}		& For example , in the 2006 World Cup in Germany, many coaches wanted teamwork .\\ \hline
\textbf{\fauxsc{amu}}		& For example , in the 2 0 0 6 world cup from Germany , as many conch wanna term work~.\\ \hline
\textbf{\fauxsc{camb}{\scriptsize 14}}		& For example , in 2006 the world cup from Germany , as many conch wanna term work~.\\ \hline
\textbf{\fauxsc{camb}{\scriptsize 16}}		& For example , in 2006 the world cup from Germany , as many conch , ' work .\\ \hline
\textbf{\fauxsc{nus}}		& For example , in 2 0 0 6 World Cup from Germany , as many conch wanna term work .	\\
\hline
\end{tabular}

\caption{\label{tab:sys-example} Examples of how human and systems corrected GUG sentences.}
\end{table*}
\section{Qualitative analysis}

We examine the system outputs of the 100 sentences analyzed in Section \ref{sec:data},
and label them by the type of errors they contain (Figure \ref{fig:sys-analysis}) and edit types made (Table \ref{tab:sys-src-edits}).
The system rankings in Table \ref{tab:gleu} correspond to the rank of systems by the percent of output sentences with errors and the percent of error-ful sentences changed. 
Humans make significantly more fluency and minimal edits than any of the systems.  The models with neural components, \nmt followed by \nus, make the most changes and produce fewer sentences with errors.  Systems often change only one or two errors in a sentence but fail to address others.  Minimal edits are the primary type of edits made by all systems (\amu and \camb made one fluency correction each, \nus two, and \nmt five) while humans use fluency edits to correct nearly 30\% of the sentences.

Spelling mistakes are often ignored: \amu corrects very few spelling errors, and even \nmt, which makes the most corrections, still ignores misspellings in 30\% of sentences.  Robust spelling correction would make a noticeable difference to output quality.
Most systems produce corrections that are meaning preserving, however, \nmt changed the meaning of 15 sentences.
This is a downside of neural models that should be considered, even though neural \mt generates the best output by all other measures.

The examples in Table \ref{tab:sys-example} illustrate some of these successes and shortcomings.
The first sentence can be corrected with minimal edits, and both \amu and \camb correct the number agreement but leave the incorrect \textit{unplanly} and the infinitival \textit{to}. In addition, \amu does not correct the spelling of \textit{advertissement} or \textit{some thing}.
\nmt changes the meaning of the sentence altogether, even though the output is fluent, and \nus makes no changes.
The next set of sentences contains many errors and requires inference and fluency rewrites to correct.
The human annotator deduces that the last clause is about coaches, not mollusks, and rewrites it grammatically given the context of the rest of the sentence.
Systems handle the second clause moderately well but are unable to correct the final clause: only \nmt attempts to correct it, but the result is nonsensical.

\section{Conclusions} 
This paper presents \corpus, a new corpus for developing and evaluating \gec systems with respect to fluency as well as grammaticality.\footnote{\url{https://github.com/keisks/jfleg}}
Our hope is that this corpus will serve as a starting point for advancing \gec beyond minimal error corrections.
We described qualitative and quantitative analysis of \corpus, and benchmarked four leading systems on this data.
The relative performance of these systems varies considerably when evaluated on a fluency corpus compared to a minimal-edit corpus, underlining the need for a new dataset for evaluating \gec.
Overall, current systems can successfully correct closed-class targets such as number agreement and prepositions errors (with incomplete coverage), but ignore many spelling mistakes and long-range context-dependent errors.
Neural methods are better than other systems at making fluency edits, but this may be at the expense of maintaining the meaning of the input.
As there is still a long way to go in approaching the performance of a human proofreader, these results and benchmark analyses help identify specific issues that \gec systems can improve in future research.

\section*{Acknowledgments}
We are grateful to Benjamin Van Durme for his support in this project.  We especially thank the following people for providing their respective system outputs on this new corpus:  Roman Grundkiewicz and Marcin Jnuczys-Dowmunt for the \amu system outputs, Mariano Felice for \camb, Zheng Yuan for \nmt, and Shamil Chollampatt and Hwee Tou Ng for \nus.  
Finally we thank the anonymous reviewers for their feedback.

\bibliography{eacl2017}
\bibliographystyle{eacl2017}

\end{document}